\documentclass{article}

\usepackage{arxiv}
\usepackage[utf8]{inputenc}
\usepackage[T1]{fontenc}
\usepackage{hyperref}
\usepackage{url}
\usepackage{booktabs}
\usepackage{array}
\usepackage{amsfonts}
\usepackage{nicefrac}
\usepackage{microtype}
\usepackage{graphicx}
\usepackage{xcolor}
\usepackage{enumitem}
\usepackage{tikz}
\usepackage{todonotes}
\usepackage{listings}
\usetikzlibrary{positioning,fit,arrows.meta,calc}
\usepackage[most]{tcolorbox}

\lstdefinestyle{promptstyle}{
    backgroundcolor=\color{gray!10},
    basicstyle=\fontsize{7}{8.5}\fontfamily{pcr}\selectfont\color{black},
    frame=single,
    rulecolor=\color{gray!50},
    frameround=tttt,
    breaklines=true,
    breakatwhitespace=true,
    columns=flexible,
    keepspaces=true,
    xleftmargin=2mm,
    xrightmargin=2mm,
    framexleftmargin=2mm,
    framexrightmargin=2mm,
    framextopmargin=1mm,
    framexbottommargin=1mm,
    upquote=true,                  
    literate={"}{\textquotedbl}1,  
}

\definecolor{claude}{HTML}{2a78d6}
\definecolor{codex}{HTML}{eb6834}
\definecolor{oracle}{HTML}{1baf7a}
\definecolor{ink2}{HTML}{52514e}

\title{nnInteractive Agent: General-Purpose AI Agents as 3D Image Annotators}

\author{
  Fabian Isensee$^{1,2}$ \quad Constantin Ulrich Harsy$^{1,3}$ \quad Maximilian Rokuss$^{1,6}$ \\
  \textbf{Benjamin Hamm$^{1,7}$ \quad Philipp Vollmuth$^{1,4}$ \quad Klaus H.\ Maier-Hein$^{1,2,3,5,6,7}$} \\[6pt]
  \parbox{\dimexpr\textwidth-2\tabcolsep}{\centering\small
  $^{1}$Division of Medical Image Computing (MIC), German Cancer Research Center (DKFZ), Heidelberg, Germany \\
  $^{2}$National Center for Tumor Diseases (NCT) Heidelberg, a partnership between the German Cancer Research
  Center (DKFZ) and Heidelberg University Hospital (UKHD), Heidelberg, Germany \\
  $^{3}$Helmholtz Imaging, German Cancer Research Center (DKFZ), Heidelberg, Germany \\
  $^{4}$Division for Computational Radiology \& Clinical AI, Clinic for Neuroradiology,
  University Hospital Bonn, Bonn, Germany \\
  $^{5}$Pattern Analysis and Learning Group, Department of Radiation Oncology, Heidelberg University
  Hospital, Heidelberg, Germany\\
  $^{6}$Faculty of Mathematics and Computer Science, Heidelberg University, Germany \\
  $^{7}$Medical Faculty, Heidelberg University, Germany \\
  }
}

\renewcommand{\shorttitle}{nnInteractive Agent}
\renewcommand{\headeright}{Technical Report}
\renewcommand{\undertitle}{Technical Report}

\begin{document}
\maketitle

\begin{abstract}
Interactive 3D segmentation models such as nnInteractive turn a few spatial prompts (points,
scribbles, boxes) into accurate volumetric masks, but a human still does the annotation work:
finding each structure, placing prompts, checking the result and correcting it. General-purpose AI
agents can now interpret images, call tools and write code. We show that such agents can take over
this work without additional training: the user makes a request in plain language, and the agent
works out what has to be segmented, locates the targets, prompts nnInteractive, inspects the 3D result, corrects it and responds
to feedback. Asked for ``the muscles used for chewing'', an agent identifies and segments the
masseters and pterygoids; asked to count liver lesions, it writes its own code to find candidates
and splits a merged lesion on request. Our open-source \verb+nninteractive-agent+ provides nnInteractive
as tools over the Model Context Protocol (MCP), together with a skill that encodes how experienced
users prompt it. On a benchmark spanning CT, MRI, PET and micro-CT of humans, a mouse and an
ant (14 images; 51 targets from organs and tumours to muscles, foot bones, vessel trees and an insect
brain), Claude Code and Codex reach a mean Dice of 0.81 and 0.84 from the request alone; when told
where each target is, both reach 0.85 and prompt nearly as effectively as an oracle that places
prompts using the reference segmentation. A text-prompted 3D model (VoxTell) is more accurate on most
standard radiological targets but less robust: it returns an empty mask for one of them and fails on
the leg muscles, the foot bones, the mouse and the ant, where the agents remain accurate. Code: \url{https://github.com/MIC-DKFZ/nninteractive-agent}
\end{abstract}

\section{Introduction}

\begin{figure}[t]
\centering
\includegraphics[width=\textwidth]{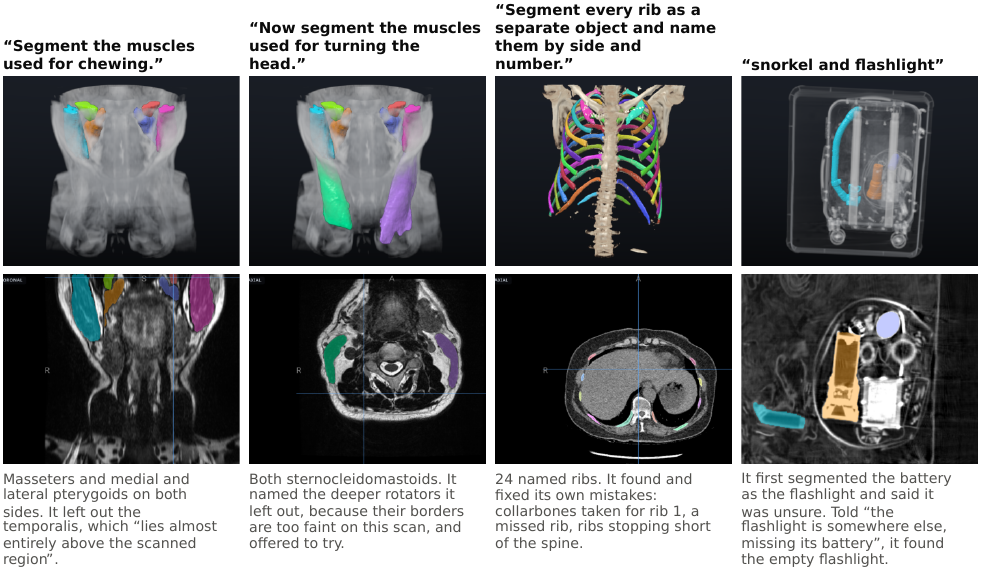}
\caption{\textbf{From a request to 3D masks.} Four requests to Claude Code (Opus 5.5 medium effort) operating
nnInteractive in the web demo: the request, the result in 3D and in one
slice (zoomed in for the suitcase), and what the agent did. The second request followed the first in the same
conversation. Images: TotalSegmentator MRI \cite{dantonoli2025totalsegmri}, RibFrac
\cite{jin2020ribfrac,yang2025ribfrac}, a CT scan of a packed suitcase (image provided by Smiths Detection).}
\label{fig:teaser}
\end{figure}

Segmentation turns images into quantitative measurements such as organ volumes, tumour burden or
cell counts. Dedicated models segment the structures they were trained on; everything else is
delineated by people, increasingly with the help of promptable models. nnInteractive
\cite{isensee2025nninteractive}, for example, predicts a full 3D mask from points, scribbles, boxes
or lassos placed in single slices and was trained on more than 120 volumetric datasets spanning CT,
MRI, PET and 3D microscopy. Such a model makes every prompt count, but the annotation process
itself remains with a human: deciding what the target is and where it lies, placing prompts,
judging the 3D result, correcting it, and revising it when someone asks for changes.

Text-prompted medical segmentation models take the human out of this loop
\cite{du2024segvol,zhao2025sat,zhao2025biomedparse,rokuss2026voxtell}. They map the name or
description of a structure directly to a mask and can be very accurate, but their accuracy
depends on how close a request is to their training data.

General-purpose AI agents such as Claude Code and Codex combine a multimodal large language model
with tool use and code execution. They interpret images, have broad anatomical knowledge, follow
multi-step instructions, write code and can check their own work. They cannot, however, produce a
voxel-accurate 3D mask by themselves. A promptable model like nnInteractive can, but it
cannot decide what to segment. The two are complementary.

Combining them, we let an agent take over the annotation process. The user makes a request in
plain language; the agent works out what has to be segmented, finds the targets, prompts nnInteractive, inspects the 3D
result, corrects it and responds to feedback, while nnInteractive turns each prompt into a 3D mask
(Fig.~\ref{fig:teaser}). Asked for the muscles used for chewing, the agent works out which muscles
these are, segments them and explains why it left one out; asked to count liver lesions, it writes
code to find candidates, flags a lesion that may be two, and splits it when the user asks
(Fig.~\ref{fig:anecdote}). No component is trained for this purpose. \emph{nninteractive-agent}
provides nnInteractive as a set of tools over the Model Context Protocol (MCP)
\cite{anthropic2024mcp}, together with a \emph{skill}, a document that describes how experienced
users prompt nnInteractive.

We evaluate nninteractive-agent with Claude Code and Codex on a benchmark of 14 images and 
compare the agents with simulated prompts and with
a text-prompted model. Agents that call segmentation models exist, including training-free agents
for 3D images (Section~\ref{sec:related}). What we add is an extension that lets an unmodified
general-purpose agent carry out the interactive 3D annotation process itself (finding, prompting,
checking, correcting) across modalities, species and structures, and an evaluation of its
segmentation accuracy. The tool, the skill, a web demo and the benchmark are open source.

\section{Related work}
\label{sec:related}

\paragraph{Promptable segmentation.} SAM \cite{kirillov2023sam} established segmentation from points
and boxes in 2D. There have also been some notable evaluations and adaptations in the medical domain~\cite{roy2023sammdzeroshotmedicalimage,wong2024scribbleprompt,ma2024medsam}. In 3D, SAM-Med3D \cite{wang2025sammed3d} and VISTA3D \cite{he2025vista3d} accept
point prompts, SegVol \cite{du2024segvol} points and boxes, and MedSAM2 \cite{ma2025medsam2} propagates a box drawn
on one slice through the volume. nnInteractive \cite{isensee2025nninteractive} accepts points, scribbles, boxes and
lassos placed in single slices and predicts a full 3D mask. In their interactive modes these models
are class-agnostic: what is segmented is decided by whoever places the prompts, usually a human.
This makes them a natural tool for an agent. We use nnInteractive for three reasons: it supports the widest range of prompt types among
these models, including scribbles, which describe elongated or irregular structures that a single
point or box cannot; it clearly outperformed existing promptable 3D methods in extensive
benchmarking \cite{isensee2025nninteractive}; and it is open-set, segmenting whatever the prompts
describe, and was trained across CT, MRI, PET and 3D microscopy.

\paragraph{Text-prompted segmentation.} SegVol \cite{du2024segvol}, SAT \cite{zhao2025sat} and
VoxTell \cite{rokuss2026voxtell} map the name or description of a structure directly to a 3D mask,
and SAM~3 \cite{carion2026sam3} does so for concepts in natural images and videos. These models
need no operator, and some are designed to generalise beyond their training data: VoxTell, our
text-prompted baseline, reports strong zero-shot results on unseen datasets and on classes related
to those it was trained on. Their accuracy on medical images still depends on how close a request is to the training
distribution, whereas that of an agent operating a class-agnostic model depends on whether it
recognises the target in the image and prompts the model well.

\paragraph{Agents that prompt segmentation models.} Closest to our work are agents that operate a
promptable model. SegAgent \cite{zhu2025segagent}, MedSAM-Agent \cite{liu2026medsamagent} and
IBISAgent \cite{jiang2026ibisagent} train multimodal models on annotation trajectories, partly with
reinforcement learning, to issue clicks and boxes to SAM-type models on 2D images. Our agent, in contrast,
is not trained at all: it learns how to prompt from the skill and keeps its general abilities. The SAM~3 Agent \cite{carion2026sam3} and the MedSAM-3 Agent
\cite{liu2025medsam3} use off-the-shelf multimodal models, but they prompt a concept segmenter with
text and inspect the returned masks on 2D images, leaving the spatial decisions to the segmenter.
Seg-Agent \cite{hao2026segagent}, also training-free, does make the spatial decisions: it places
visual prompts for SAM and refines them after inspecting candidate masks drawn on 2D natural images.
In MedOpenClaw \cite{shen2026medopenclaw}, frontier agents operate 3D~Slicer and can call VISTA3D,
but segmentation is an auxiliary step for diagnostic questions, its accuracy is not evaluated, and
the authors found the agents' multi-step segmentation workflows unreliable. In our work, interactive 3D
segmentation is the task itself: the agent places spatial prompts, including scribbles, checks the
result in rendered slices and 3D views, and corrects it. We evaluate this process for segmentation accuracy across imaging modalities
and species.

\paragraph{Training-free agents and imaging software.} Agents built on frontier models without
additional training are increasingly used in medical imaging, also for 3D images. Erdur et al.\
\cite{erdur2026agentic} run brain MRI pipelines (registration, automatic tumour and anatomy
segmentation, volumetry) through text-only tool calls, and CT-Flow \cite{gu2026ctflow} provides CT
viewing, measurement and automatic segmentation tools through MCP servers for question answering
and reporting. HealthAgentBench \cite{liu2026healthagentbench} evaluates unmodified Claude Code and
Codex on healthcare tasks and finds imaging tasks, including 3D CT, the hardest, which suggests that
agents need suitable tools for images. Agents can also operate imaging software directly, by writing
code in napari \cite{royer2024omega} or through generic MCP servers for 3D~Slicer and napari
\cite{mcpslicer2025,naparimcp2025}. Most of these systems rely on automatic segmentation models,
which limits them to the structures those models know, and many are dedicated agent frameworks
around their own tools or viewers. nninteractive-agent instead extends an existing general-purpose
agent: an MCP server and a skill that any Claude Code or Codex session can load, without a viewer.

\section{nninteractive-agent}
\label{sec:system}

\begin{figure}[t]
\centering
\includegraphics[width=\textwidth]{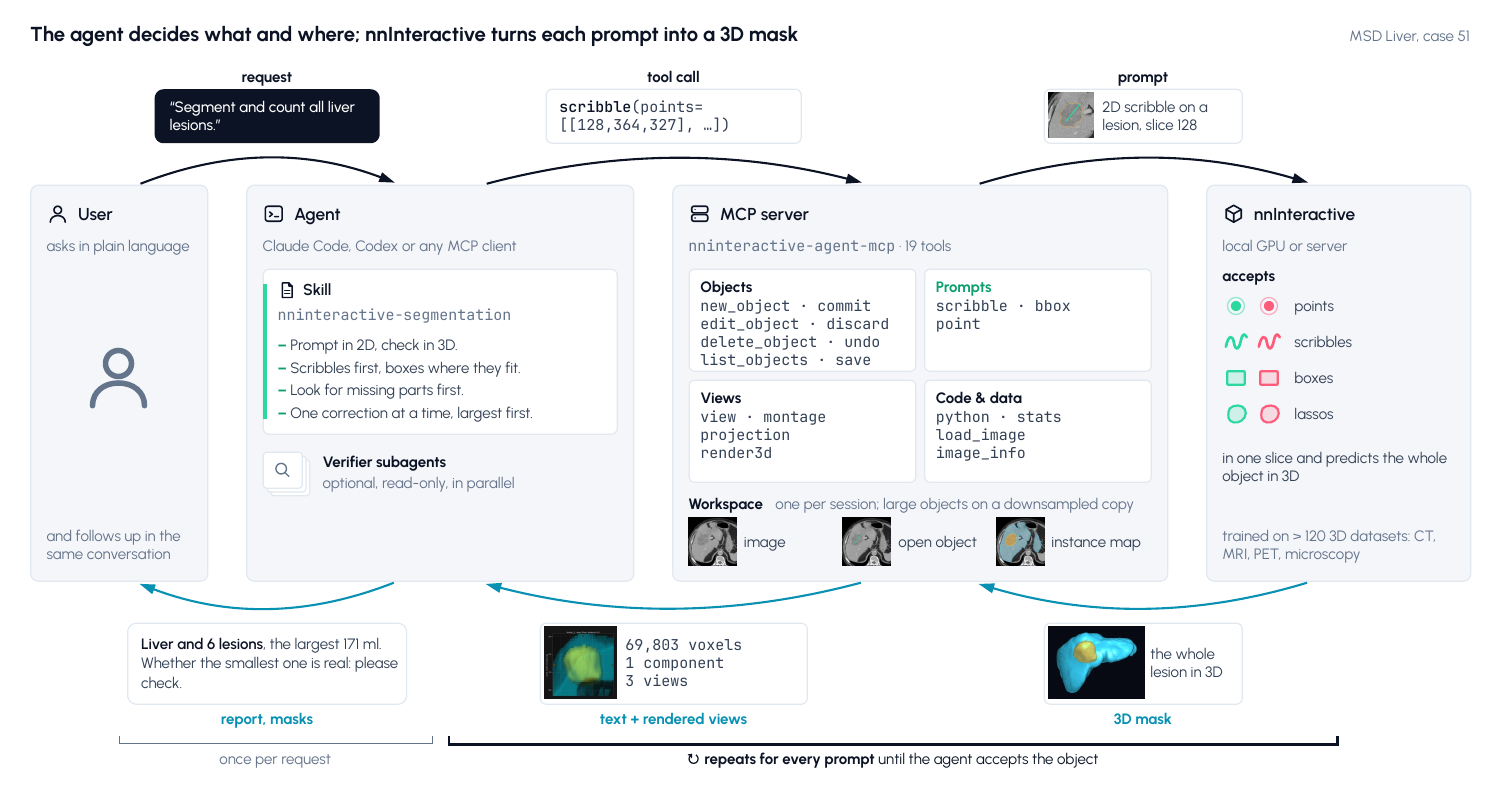}
\caption{\textbf{Overview.} The agent reads the skill and works through the tools of the MCP
server. The server holds the image, the objects segmented so far and an nnInteractive session.
Each prompt goes through the server to nnInteractive; the agent checks the resulting 3D mask in the
returned text and in rendered views, and repeats this until it accepts the object. It reports to the
user once per request. Example: ``Segment and count all liver lesions'' on MSD Liver, case 51
\cite{simpson2019msd,antonelli2022msd}. nnInteractive runs locally or on a shared server.}
\label{fig:architecture}
\end{figure}

\subsection{Division of labour}

The agent decides \emph{what} to segment and \emph{where} to prompt, and it judges the result;
nnInteractive turns each prompt into a 3D mask (Fig.~\ref{fig:architecture}). The agent never
draws a mask itself and never needs to be precise to the voxel: a scribble along the middle of a
structure is enough for nnInteractive to find its boundary in 3D. Conversely, nnInteractive needs
no knowledge of what a ``left parotid gland'' is: the agent supplies it through the location of
its prompts. This split puts each capability where it is strongest, and nothing in the system is
specific to an anatomy, a modality or a species.

\subsection{Tools and workspace}

The MCP server keeps one \emph{workspace} per agent session: the image, one nnInteractive session,
the object being segmented with its prompts, and an instance map of the objects committed so far.
The agent works with 19 tools in four groups.

\paragraph{Objects.} nnInteractive segments one object at a time. The agent starts an object, adds
prompts, and commits it to the instance map, where every voxel belongs to one object. A structure
inside another one, such as a lesion in an organ, takes those voxels from the outer object when
the agent asks for it; otherwise an overlap with a committed neighbour is treated as a leak and
reported. Committed objects can be reopened for corrections.

\paragraph{Prompts.} A \emph{point} marks a tiny object or a tiny error; a \emph{box} encloses the
object in one slice and is allowed only as an object's first prompt; a \emph{scribble} is a line
drawn in one slice. nnInteractive accepts any binary mask as a scribble; the agent specifies a
polyline, which the tool rasterises 2 pixels wide. Points and scribbles can be positive or
negative. Boxes and an object's first scribble are previewed: the tool draws them on a zoomed
view without sending them, so the agent can check the coordinates it read off a rendered image. After every prompt the tool reports how the mask
changed and flags common problems, such as a prediction that extends past the box it came from.

\paragraph{Views.} The agent sees the image only through rendered views
(Fig.~\ref{fig:agentview3d}; more examples in Figs.~\ref{fig:sup_heart}--\ref{fig:sup_ant} of the
supplement): orthogonal slices with absolute voxel coordinates on the axes,
patient directions and object outlines; montages of many slices; projections; and 3D surface
renderings of an object from several directions. Statistics (size, connected components,
overlaps) complement them.

\paragraph{Python.} A persistent Python namespace gives the agent the image array and the workspace.
The agent uses it to find candidates (thresholds, connected components, maximum intensity
projections), to loop over many instances, and to draw its own figures (Fig.~\ref{fig:anecdote}a).
The tool can be restricted to computations or switched off.

Objects larger than about 320 voxels along an axis are segmented on a downsampled copy of the
image and scaled back up, since nnInteractive predicts 192$^3$ patches and is most accurate on
objects that fit into one patch or require only small AutoZoom factors. The agent only provides a
size estimate and continues to work in full-resolution coordinates.

\begin{figure}[t]
\centering
\includegraphics[width=\textwidth]{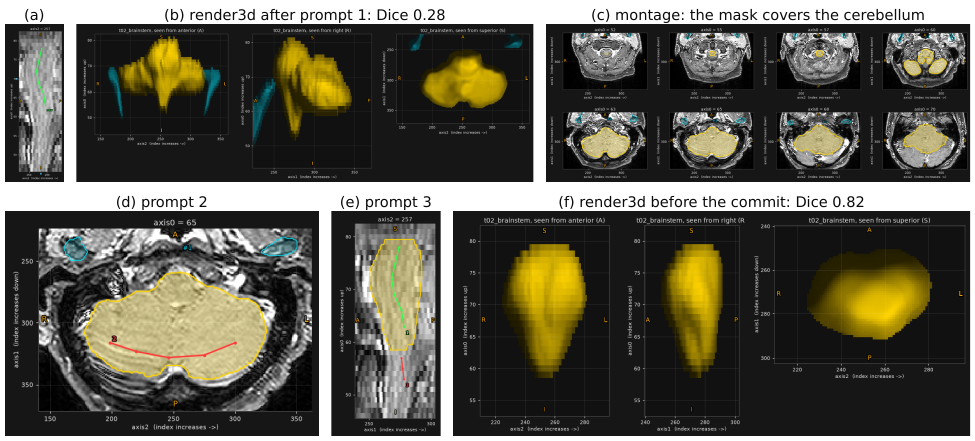}
\caption{\textbf{What the agent sees}, here Codex segmenting the brainstem in a head-and-neck MRI
(\texttt{headneck\_mr}, HaN-Seg \cite{podobnik2023hanseg}; unguided track). (a) The preview of its
first scribble, in a sagittal slice. (b) The 3D rendering after this prompt: the object (169~ml)
extends far to both sides (Dice 0.28), and (c) the montage shows that it covers the cerebellum.
(d) The second prompt, a negative scribble through the cerebellum in an axial slice, removed most of
the leak (Dice 0.80). (e) The third prompt, a short negative scribble at the lower end, shown in a
sagittal slice with the mask after the second prompt. (f) The 3D rendering before the commit (Dice
0.82). The agent committed the brainstem with the note ``Brainstem separated from cerebellum and
cervical cord; superior midbrain boundary and cervicomedullary transition are approximate on 3~mm
slices.'' Panels (a)--(c) and (f) are renders the agent looked at. It sent the two corrections
without a preview, so (d) and (e) were drawn afterwards with the same renderer from the logged
prompts and masks.}
\label{fig:agentview3d}
\end{figure}

\subsection{The skill}

Tools alone do not make a good annotator. The skill, a Markdown document that agents load when
asked to segment, encodes how experienced users work with nnInteractive. It describes how
nnInteractive behaves, how to read coordinates off rendered views, a step-by-step workflow, a
playbook for corrections and rules for working efficiently (Fig.~\ref{fig:skill}). Skills of this
kind have also been shown to help agents operate scientific visualisation software
\cite{ai2026scivisagentskills}.

\begin{figure}[t]
\begin{lstlisting}[style=promptstyle]
- One object at a time. Each object is a single structure instance. Segmenting "the rib
  cage" means one object per rib [...]
- 2D prompts, 3D result. A prompt lives in one slice, but the prediction is a full 3D
  mask. Always check slices other than the one you prompted in.
- Prompts accumulate. [...] `undo` it right away if a prompt made things worse.
- Prompt types: scribbles preferred, boxes only where they fit, points rarely.
- Long lines work best: run the line along the middle of the region it describes and let
  it cover most of the region's visible extent [...]
- You decide what belongs to the object. [...] state the decision in your report.
- Look for missing parts first. A result that looks clean is often too small [...]
- Correct the errors from largest to smallest, one prompt at a time, re-checking after
  each [...]
- To check more than about 4 objects, split them into batches of 3-6. Verifier subagents
  can check the batches in parallel [...]
- [...] summarize what was segmented, how confident you are per object, and anything a
  human should review.
\end{lstlisting}
\vspace{-4pt}
\caption{\textbf{Excerpts from the skill} (verbatim; [...] marks omitted text).}
\label{fig:skill}
\end{figure}

\subsection{Deployment}
\label{sec:deployment}

\texttt{nninteractive-agent setup claude} (or \texttt{codex}) registers the MCP server and installs
the skill, so that every new session can segment images; any other MCP client can use the server,
and clients without skill support read the skill as an MCP prompt. nnInteractive runs in the
server's process on a local GPU or on a shared \texttt{nninteractive-server}. Open models behind
an Anthropic-compatible API can drive Claude Code as well. As a proof of concept, Qwen3.8-27B segmented
a spleen (MSD Spleen, case 10 \cite{simpson2019msd,antonelli2022msd}) with Dice 0.93 in 34 minutes,
and the benchmark's bladder tumour, with the location given, with Dice 0.60 (Claude Code and Codex:
0.72 to 0.81). It ran on a shared server under high load, so we did not pursue this further. A web demo with 2D and 3D views and a chat lets people use the agent with immediate visual feedback (Fig.~\ref{fig:webdemo} in the supplement).

\section{Agentic behaviour in practice}
\label{sec:examples}

Before we measure how accurately agents segment named targets (Section~\ref{sec:benchmark}), we
show what an agent does with requests as people make them: often less specific, sometimes defined
by function rather than by name, and with feedback along the way. The following examples come
from the web demo with Claude Code (Opus 5.5, medium effort).

\begin{figure}[t]
\centering
\includegraphics[width=\textwidth]{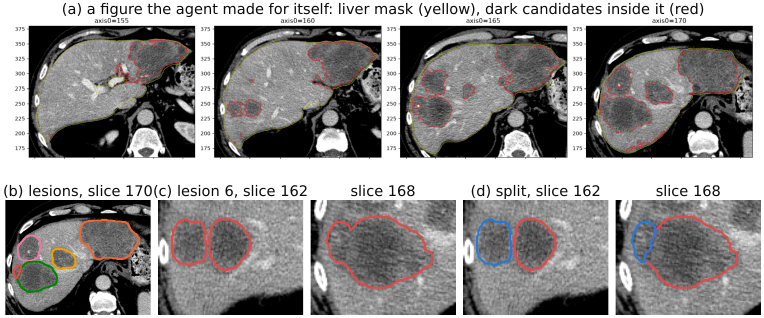}
\caption{\textbf{Liver lesions from a one-line request}. (a) A figure the agent drew for itself
with its Python tool to review lesion candidates. (b) The committed lesions in one slice.
(c) Lesion~6 as first segmented and (d) after the user asked to split it.}
\label{fig:anecdote}
\end{figure}

\subsection{Finding and counting liver lesions}
\label{sec:lesions}

Asked only to ``segment and count all liver lesions'' in a contrast-enhanced CT (MSD Liver,
case 155 \cite{simpson2019msd,antonelli2022msd}), the agent made its own plan
(Fig.~\ref{fig:anecdote}, box below). It first segmented the liver as a helper object. In
Python, it smoothed the image, thresholded dark voxels inside the eroded liver mask, labelled the
connected components and drew a figure of candidate regions for itself
(Fig.~\ref{fig:anecdote}a). It then segmented each lesion as its own nnInteractive object, mostly
starting from a box, separated two touching nodules with negative scribbles, reviewed the
remaining candidates (dismissing vessels, fissures and noise), and deleted the helper. After three
minutes and ten prompts it reported six definite lesions and one possible one, and flagged that
lesion~6 might be two touching nodules. When told ``lesion 6 seems to be a merger. Split into 2'', it
checked the split with an intensity-based watershed of its own, reopened the object, removed the
lateral nodule with a negative scribble, segmented that nodule as a new object, and confirmed that
both boundaries agreed (Fig.~\ref{fig:anecdote}c,d); this took 46 seconds and two prompts.

\begin{tcolorbox}[enhanced, colback=white, colframe=ink2!50, boxrule=0.4pt, arc=2pt, left=4pt, right=4pt, top=2pt, bottom=2pt, fontupper=\small, title={Conversation excerpt (shortened)}, coltitle=black, colbacktitle=ink2!10, fonttitle=\small\bfseries]
\textbf{User:} segment and count all liver lesions\\
\textbf{Agent:} I've got a liver mask; now finding hypodense lesion candidates inside it. [\dots]
\emph{I count 6 definite liver lesions and 1 possible one.} [\dots] To find them, I outlined the
liver as a temporary helper (deleted afterwards), searched it for dark regions, and looked at each
candidate in zoomed slices. [\dots] \emph{Whether \#6 is one lesion or two:} at axis0 158--162 it
shows as two touching nodules that merge higher up. I kept it as one mass.\\
\textbf{User:} lesion 6 seems to be a merger. Split into 2\\
\textbf{Agent:} Splitting \#6 into a lateral and a medial nodule. These are clearly separate at
axis0 158--162. [\dots] The split agrees with a separate check based on image brightness, which put
the boundary in the same place. \emph{Count: 7 definite lesions (\#1--6, \#8), plus the uncertain
small spot \#7.}
\end{tcolorbox}

\subsection{Requests that need knowledge}
\label{sec:knowledge}

Four further requests (Fig.~\ref{fig:teaser}) need knowledge rather than search. The first three
need anatomical knowledge; the fourth is not a medical image at all:
\begin{itemize}[leftmargin=1.5em,itemsep=1pt,topsep=2pt]
  \item \textbf{``Segment the muscles used for chewing.''} (TotalSegmentator MRI, s0506
    \cite{dantonoli2025totalsegmri}): the agent translated function into anatomy and segmented
    both masseters and both medial and lateral pterygoids (2 minutes 18 seconds, 7 prompts). It
    left out the temporalis and explained why: it ``lies almost entirely above the scanned region''.
  \item \textbf{``Now segment the muscles used for turning the head.''}: both sternocleidomastoids
    (1 minute 30 seconds, 2 prompts). The agent named the deeper rotators it did not segment,
    because their borders were too faint on this scan, and offered to try.
  \item \textbf{``Segment every rib as a separate object and name them by side and number.''}
    (RibFrac, case 100 \cite{jin2020ribfrac,yang2025ribfrac}): the agent found 12 rib candidates per side with its
    own bone thresholding and prompted each with points along it. It noticed that its first pass
    had taken both collarbones for rib~1 and missed the right fifth rib under the shoulder blade,
    and its batch-wise reviews found that most ribs stopped short of the spine. It fixed all of
    this, rebuilt three ribs whose extensions had spilled into the spine, and reported that the
    numbering was counted from the first rib and not checked against the vertebrae. The result is
    24 named ribs after 10.6 minutes and 223 prompts (\$3.86 at API prices).
  \item \textbf{``snorkel and flashlight''} (a CT scan of a packed suitcase, image provided by Smiths Detection): an image
    far from the medical and microscopy images nnInteractive was trained on. The agent recognised the
    snorkel as the J-shaped tube along one side of the case and segmented it, including its
    mouthpiece, with four scribbles. For the flashlight it chose a dense metal cylinder after ruling
    out several other items, and said it was unsure: it saw no lens, reflector or switch, and the
    cylinder could also be a large battery (3 minutes 12 seconds, 5 prompts). Told that this was
    likely the battery and that the flashlight, missing its battery, was somewhere else, the agent
    found an empty tube with a flared head next to it, which it had taken for a cup-like container
    before. It segmented the tube with one scribble and re-created the first object as ``battery''
    (46 seconds, 2 prompts).
\end{itemize}

\section{Benchmark}
\label{sec:benchmark}

\begin{figure}[t]
\centering
\includegraphics[width=\textwidth]{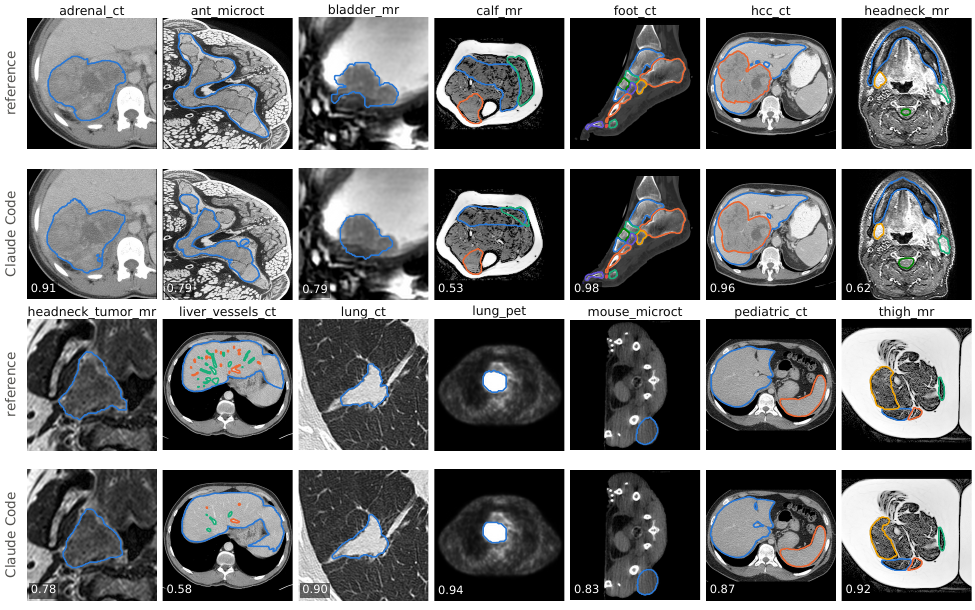}
\caption{\textbf{The benchmark images}, one slice each. Upper rows: the reference; lower rows: what
Claude Code segmented from the request alone (unguided track, the middle of three repeats), with
that run's Dice averaged over the image's targets. Fat-weighted MRI (calf, thigh) is shown on a log
scale; for the whole-body CT (\texttt{foot\_ct}), a sagittal slice through the left foot in a bone
window. All tracks: Figs.~\ref{fig:casesall1} and \ref{fig:casesall2} in the supplement.}
\label{fig:cases}
\end{figure}

\paragraph{Images and targets.} The benchmark consists of 14 public images: six CT, five MRI,
one PET and two micro-CT, of humans, a mouse and an ant (Fig.~\ref{fig:cases},
Table~\ref{tab:sources}). Its 51 targets range from organs and tumours to muscles, bones, vessel
trees and an insect brain. Every target is identifiable from its description alone: it is named,
unique within the image, and its side is specified where relevant. Besides all 14 images, we report
two groups: the nine images of human CT, MRI and PET of the head and neck, chest, abdomen and pelvis
(\emph{radiology}; 25 targets), the typical domain of medical segmentation models, and five images
whose targets lie outside it (\emph{legs, mouse, ant}; 26 targets): single muscles of the thigh and
calf in fat-weighted research MRI, the 17 bones of the left foot in a whole-body CT of a cadaver,
and a tumour and a brain in micro-CTs of a mouse and an ant.
nnInteractive was not trained on any of these images, so the agents' results are not inflated by
training-data overlap; several sources are part of VoxTell's training data
(Section~\ref{sec:voxtell}). The references of the foot were made by the source's annotators with
nnInteractive in 3D~Slicer and checked by them, which may favour the tracks that use nnInteractive.

\paragraph{Tracks.}
\begin{itemize}[leftmargin=1.5em,itemsep=1pt,topsep=2pt]
  \item \textbf{Unguided}: the agent gets the image and the list of targets by name
    (Fig.~\ref{fig:requests} in the supplement). It has to find and segment them.
  \item \textbf{Guided}: the same, plus one voxel inside each target (the voxel deepest inside the
    reference). This separates segmenting from finding.
  \item \textbf{One point}: no agent; one positive point at the guided voxel per target. This is
    nnInteractive with the simplest possible prompt.
  \item \textbf{Oracles}: no agent; up to five simulated prompts per target with knowledge of the
    reference, each aimed at the largest error of the current mask (the largest connected region
    it misses or adds; before the first prompt, the whole target). The \emph{point oracle} places
    a point at the voxel deepest inside that region, the \emph{scribble oracle} a line along the
    middle of the region (the longest path through its skeleton) in the slice where the region is
    largest. Both use the same tools as the agents. We record the Dice after every prompt. As the
    agents prompt almost only with scribbles (Table~\ref{tab:behaviour}), we compare each agent with
    the scribble oracle at the number of prompts the agent sent on each target (after five prompts for the 4--11\% of targets where an agent sent more).
  \item \textbf{VoxTell}: no agent; VoxTell v1.1 \cite{rokuss2026voxtell} with exactly the target
    texts the agents get, all targets of an image in one call (for the 17 foot bones three per call,
    since VoxTell's outputs for all 17 do not fit into the memory of our GPU; on another image,
    splitting the prompts changed its masks by less than 0.002 Dice).
\end{itemize}

\paragraph{Scoring.} Each target is scored by Dice, using its object as it was when the agent last
committed it; a later object that takes voxels from it (a tumour inside the liver) does not count
against it. A missing target scores 0. Means are taken per image first: the targets of an image are
averaged, and a track's mean is the mean over the images (all 14, or those of one group), so that the
image with 17 targets counts as much as an image with one.

\paragraph{Setup.} nnInteractive ran locally with its public v1.0 weights on one NVIDIA RTX 4090.
The agents were Claude Code 2.1 with Claude Opus 5.5 and Codex 0.162 with gpt-6.1-sol, both at
medium reasoning effort. They worked through the nnInteractive tools, including a Python tool for
their own code.
Each agent track and the one-point track were run three times; the oracles and VoxTell once.

\section{Results}
\label{sec:results}

\begin{table}[t]
\centering
\caption{\textbf{Benchmark results} (mean Dice over images, each image's targets averaged first;
agents: mean $\pm$ standard deviation of the track mean over three repeats, i.e.\ run-to-run
variation; for the uncertainty over images see Section~\ref{sec:click}). Numbers of images in
brackets. Radiology: human CT, MRI and PET of the head and neck, chest, abdomen and pelvis (25
targets); legs, mouse, ant: the thigh and calf MRIs, the foot bones in a whole-body CT, and the
micro-CTs of a mouse and an ant (26 targets). Rad.\ without \texttt{adrenal\_ct}: the radiology images
except the one on which VoxTell returns an empty mask (Section~\ref{sec:voxtell}). ``Scribble oracle, same prompts'': its Dice after as
many prompts on each target as the agent sent on it (after five where the agent sent more). Best
Dice per column in bold.}
\label{tab:results}
\small
\setlength{\tabcolsep}{3.4pt}
\begin{tabular}{lccccccc}
\toprule
Track & \multicolumn{4}{c}{Dice} & Prompts & Scribble oracle, & Minutes \\
\cmidrule(lr){2-5}
 & all (14) & radiology (9) & rad. without & legs, mouse, & per target & same prompts & per image \\
 &          &               & \texttt{adrenal\_ct} (8) & ant (5) & & & \\
\midrule
Codex, guided          & \textbf{0.855} $\pm$ 0.002 & 0.823 & 0.812 & \textbf{0.912} & 2.9 & 0.869 & 3.8 \\
Claude Code, guided    & 0.847 $\pm$ 0.006 & \textbf{0.828} & 0.818 & 0.880 & 2.2 & 0.844 & 3.6 \\
Codex, unguided        & 0.842 $\pm$ 0.004 & 0.815 & 0.804 & 0.892 & 2.8 & 0.886 & 4.2 \\
Claude Code, unguided  & 0.814 $\pm$ 0.006 & 0.814 & 0.803 & 0.813 & 2.3 & 0.854 & 4.3 \\
\midrule
One point, no agent    & 0.709 & 0.669 & 0.651 & 0.779 & 1   & 0.769 & 0.2 \\
VoxTell, text only     & 0.539 & 0.777 & \textbf{0.875} & 0.110 & --  &       & 1.5 \\
\bottomrule
\end{tabular}
\end{table}

\subsection{The agents' prompts are far better than a single click}
\label{sec:click}

With the request alone, the agents reach a mean Dice of 0.81 (Claude Code) and 0.84 (Codex); with
one voxel per target as a hint, both reach 0.85 (Table~\ref{tab:results}). nnInteractive with one
click per target at that same voxel reaches 0.71. The gain is largest where a single point
cannot describe the target: on the liver of \texttt{liver\_vessels\_ct}, one point gives Dice 0.00
and the agents 0.95 to 0.96; on the liver tumour of \texttt{hcc\_ct}, 0.13 against 0.96. Locating the
targets costs the agents little: the unguided tracks are only 0.01 (Codex) and 0.03 (Claude Code)
below the guided ones. No agent run left a target without an object.

With 14 images, these means carry considerable uncertainty: 95\% intervals from a bootstrap over
images are 0.11 to 0.14 wide for the agents, for example [0.79, 0.91] for Codex, guided.
Comparisons on the same images are more precise. Every agent track is ahead of one click, by
0.11 [0.01, 0.22] (Claude Code, unguided) to 0.15 [0.05, 0.26] (Codex, guided). The location hint
helps Claude Code (+0.03 [0.01, 0.07]) but not Codex significantly (+0.01 [$-$0.001, 0.027]). The
two agents do not differ clearly: without the hint, Codex is ahead by 0.03, but the interval includes
zero ([$-$0.003, 0.070]); with the hint, the difference is 0.01 [$-$0.01, 0.03]. The Dice of every
target in every track is listed in Table~\ref{tab:pertarget} of the supplement.

\subsection{The agents' prompts are close to the oracles'; they stop earlier}

\begin{figure}[t]
\centering
\includegraphics[width=\textwidth]{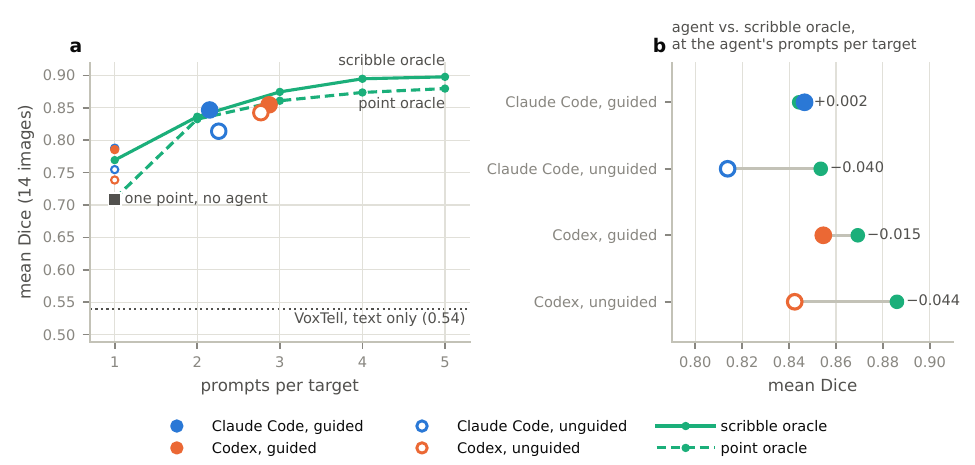}
\caption{\textbf{Agents compared with oracles} (mean Dice over images). (a) Dice over the number of
prompts per target: oracle trajectories, the agents at their mean number of prompts (filled:
guided, hollow: unguided) and after their first prompt on each target (small markers at one
prompt), nnInteractive with one point, and VoxTell. (b) Each agent next to the scribble oracle given
the agent's own number of prompts on each target.}
\label{fig:results}
\end{figure}

To judge the quality of the agents' prompts, we compare them with the scribble oracle at the same number of prompts:
for every target, its Dice after as many prompts as the agent sent on it
(Table~\ref{tab:results}, Fig.~\ref{fig:results}b). With the location given, the scribble oracle
is level with Claude Code (0.844 against 0.847; 95\% interval of the difference [$-$0.03, 0.02]) and
0.015 ahead of Codex ([$-$0.01, 0.04]). Without the location, the scribble oracle is further ahead
(0.04 for both agents): it knows where each target lies; the agents have to find it. After their
first prompt, the guided agents reach 0.79, about as much as the scribble oracle after one prompt
(0.77; the agents see their first scribble or box in a preview before sending it, whereas the
oracle's first scribble follows a fixed rule), and without the hint 0.74 (Codex) and 0.76 (Claude
Code). The agents send 2.2 to 2.9 prompts per target,
while the oracles keep improving with every further prompt (Fig.~\ref{fig:results}a): the agents
judge their results from rendered views and stop when a result looks right, whereas the oracles
compare each result with the reference, whose exact extent a request in plain language cannot fully
convey.

\subsection{A text-prompted model is more accurate within its domain, but brittle}
\label{sec:voxtell}

VoxTell segments all targets of an image in about 20 seconds from the target texts alone (17
minutes for the 17 foot bones in the whole-body CT) and reaches a mean Dice of 0.54. This average
conceals two distinct regimes. On seven of the nine radiology images (and on 19 of their 25 targets)
it is more accurate than every agent, by 0.02 to 0.09 per image, and it does well on thin structures
that the agents find hard (left optic nerve 0.59 against 0.02 to 0.34 for the agents, spinal cord
0.61 against 0.46 to 0.55, hepatic veins 0.64 against 0.27 to 0.53); on \texttt{hcc\_ct} it is
level with them. On the adrenal tumour, however,
it returns an empty mask, so its mean over the nine images (0.78) is below the agents' (0.81 to 0.83;
Table~\ref{tab:results}); without this image, it is 0.88 against 0.80 to 0.82. The empty mask is a
matter of wording: ``the tumor of the right adrenal gland'', the benchmark's request, gives Dice 0,
while ``adrenal tumor'' gives 0.79 (a check outside the benchmark). Several of the radiology source
datasets are part of VoxTell's training data (HaN-Seg, HNTS-MRG, RIDER Lung CT, HCC-TACE-Seg), so
these numbers may be optimistic. On the second group VoxTell fails: it reaches 0.10 on the leg
muscles (agents 0.74 to 0.87), 0.03 on the foot bones (agents 0.98), where five of its 17 masks are
empty and most of the others mainly cover a different bone, and 0.17 on the mouse and the ant (agents
0.80 to 0.92). Its vocabulary of precomputed prompts names all of these muscles and each tarsal bone
(the metatarsals and toe bones only as groups); the images, however, are far from standard radiology:
fat-weighted research MRI, a whole-body CT of a cadaver and animals. This is the
expected behaviour of a model that maps text to masks directly. The agent, in contrast, does not map text to masks: it recognises the target in the image and
prompts nnInteractive.

\subsection{Where the agents struggle}

Thin and tubular structures are the hardest targets for the agents: the portal and hepatic veins
(0.27 to 0.53), the left optic nerve (0.02 to 0.34) and the spinal cord (0.46 to 0.55). They are
hard to prompt, since a scribble must follow a narrow structure across slices, and hard to judge from renderings. Per-image results for all tracks are
in Fig.~\ref{fig:perimage} and Figs.~\ref{fig:casesall1}--\ref{fig:casesall2} of the supplement,
per-target results in Table~\ref{tab:pertarget}.
Without the location hint, Codex is ahead of Claude Code (0.84 against 0.81), mostly on the calf
muscles and the ant brain; on this small benchmark, the difference is not clearly significant
(Section~\ref{sec:click}).

\subsection{How the agents work}

\begin{table}[t]
\centering
\caption{\textbf{How the agents work} (all runs: 14 images, three repeats each). Prompt types as a share of all
prompts sent; ``from code'': prompts the agent sent from its own Python code. Prompts and views per
target: means over images.}
\label{tab:behaviour}
\small
\begin{tabular}{lcccccc}
\toprule
 & Prompts per target & Scribbles & Boxes & Points & From code & Views per target \\
\midrule
Claude Code, guided    & 2.2 & 94\%  & 4\% & 2\% & 14\% & 9.1 \\
Claude Code, unguided  & 2.3 & 94\%  & 3\% & 2\% & 17\% & 10.4 \\
Codex, guided          & 2.9 & 100\% & 0\% & 0\% & 0\%  & 13.9 \\
Codex, unguided        & 2.8 & 99\%  & 0\% & 1\% & 0\%  & 14.3 \\
\bottomrule
\end{tabular}
\end{table}

Both agents follow the skill: almost all prompts are scribbles, boxes and points are rare
(Table~\ref{tab:behaviour}). Claude Code sends about a sixth of its prompts from its own Python
code; Codex prompts only through the tools but looks at more views, including four to six times as
many 3D renderings. An image takes about four minutes (3.6 targets on average). At API prices,
Claude Code's runs cost \$1.29 (guided) and \$1.46 (unguided) per image (\$115 for both tracks
together), mostly for 2.8 to 3.1 million cached input tokens; Codex used about 1.7 to 1.9 million
input tokens per image, 96\% of them cached. The agents' notes are informative: objects whose commit
note expresses uncertainty (``approximate'', ``uncertain'', \dots) reach a mean Dice of 0.72 (Claude
Code) and 0.68 (Codex), against 0.92 and 0.93 for the others.

\section{Discussion}

\paragraph{An extension, not a new agent.} nninteractive-agent adds a capability to existing agents
rather than building a new one. This has three consequences. First, it is not tied to one model: the
same tools already serve two commercial agents and an open model, and we expect a better model to
become a better annotator without any change on our side.
Second, the agent keeps its general abilities (anatomical knowledge, coding, web search,
conversation), on which the examples in Section~\ref{sec:examples} rely and which a specialized model
fine-tuned for promptable segmentation tasks does not provide to the same extent. Third, segmentation
becomes one tool among others in the agent's usual environment: it needs no dedicated viewer, its
results can feed into whatever else the agent does with the data, and further tools, such as a
text-prompted model, a detector or a registration tool, could be added in the same way. The other side of this is a dependence on the agents' providers, both for the agents'
quality and for which data may be sent to them (see Limitations).

\paragraph{What nninteractive-agent enables.} With nninteractive-agent, an existing agent takes over
the part of interactive segmentation that a promptable model leaves to its user. Told where a target is, it prompts nnInteractive almost as
effectively as a simulated user who knows the answer, at the same number of prompts, and without
that hint it loses little. More importantly, it handles requests that no
fixed interface anticipates: counting lesions it first has to find, muscles described by their
function, two dozen ribs that have to be numbered. It brings anatomical knowledge, plans the work,
writes code where code is the better tool, checks its results, and says what it is unsure about.
These abilities are the agent's own; nninteractive-agent, a set of tools and a skill, connects them
to nnInteractive without training a model.

\paragraph{Agentic versus text-prompted segmentation.} The comparison with VoxTell shows the
trade-off. A text-prompted model is fast and, on most standard radiology targets, more accurate than
the agents. It is also brittle: a request worded differently from what it expects can return an
empty mask, and on images far from standard radiology (single muscles in fat-weighted MRI, the bones
of the foot, animals) it fails. The agent needs to recognise the target in the image
and to prompt nnInteractive, which is not tied to a fixed set of structures; it therefore keeps working where
the text-prompted model stops, at the price of minutes per image. The two are complementary. A
text-prompted model could be one more tool for the agent, giving a first draft for standard
targets that the agent checks and corrects with nnInteractive; we leave this for future work.

\paragraph{Limitations.} The agents stop earlier than an oracle would; better self-assessment is
the most direct way towards oracle performance. Thin and tubular structures remain difficult to
prompt and to judge from renderings. A run takes minutes and, at API prices, costs \$1.3 to \$1.5
per image, which suits annotation and research use better than the processing of large cohorts.
We did not compare the agents with human annotators. An experienced nnInteractive user would
likely be more accurate, particularly on thin structures, and faster per target; the agent's
advantage is that nobody needs to sit at the screen. With a commercial agent, rendered views of the image are sent to the LLM provider, which has to be
compatible with the terms of use of the data; open models served in-house avoid this; we tested one
only as a proof of concept (Section~\ref{sec:deployment}). The benchmark is small (14 images, 51 targets), the oracles
follow simple rules and show what well-placed prompts achieve rather than an upper bound. The nnInteractive and VoxTell weights are licensed for
non-commercial use only.

\section{Conclusion}

General-purpose AI agents can act as 3D image annotators: with a set of tools and a skill, and
without additional training, they operate nnInteractive much as human annotators do. They place prompts in slices, check the 3D result and correct it. From
plain-language requests they reach a mean Dice of 0.81 to 0.84 on a diverse benchmark, and
0.85 when told where each target is, prompting nearly as effectively as an oracle given the same
number of prompts; they keep working on data where a text-prompted model fails. Beyond prompting, they plan,
write their own code and take corrections in plain language. As agents
improve, we expect this way of annotating images to improve with them.

\paragraph{Code availability.} nninteractive-agent, comprising the MCP server, the skill, the web
demo, the benchmark scripts and the aggregated benchmark results (per track and per image):
\url{https://github.com/MIC-DKFZ/nninteractive-agent} (Apache-2.0). nnInteractive:
\url{https://github.com/MIC-DKFZ/nnInteractive}.

\paragraph{Data availability.} All images used in this work, except the suitcase CT scan in
Fig.~\ref{fig:teaser} (provided by Smiths Detection), are available from their original sources, the thigh and calf MRI on request; we do not redistribute any image or reference mask. The benchmark sources, their
licences and references are listed in Table~\ref{tab:sources}. The build script
(\texttt{benchmark/prepare.py}) downloads and converts them, derives the reference masks as defined
in \texttt{benchmark/cases.json} and verifies the result against published checksums. Some sources
require a free registration (VEELA) or a manual download in a web browser (ant brains). The thigh
and calf MRI is no longer in its figshare record and is available from its authors on
request.\footnote{\url{https://github.com/VinceWohl/DL-Segmentation-of-Leg-Muscles}, which names the
contact.} The Visible Human images are courtesy of the U.S. National Library of Medicine. The examples in
Section~\ref{sec:examples} use MSD Liver (cases 51 and 155; CC BY-SA 4.0)
\cite{simpson2019msd,antonelli2022msd}, TotalSegmentator MRI v2.0.0 (case s0506; CC BY-NC-SA 2.0)
\cite{dantonoli2025totalsegmri} and RibFrac (case 100; CC BY-NC 4.0)
\cite{jin2020ribfrac,yang2025ribfrac}; the open-model test in Section~\ref{sec:deployment} uses MSD
Spleen (case 10; CC BY-SA 4.0).

\paragraph{Acknowledgements.} This work was supported by the NCT Heidelberg Image Data Analysis
Infrastructure (IDAI) and by the Helmholtz Association under the Helmholtz Foundation Model Initiative (HFMI), project “The Human Radiome Project” (THRP). MR was supported by the Google PhD Fellowship Program. BH was funded by
``NUM 3.0'' (FKZ: 01KX2524) and by the Vector Stiftung. PV is partially funded by an Else Kr\"oner
Clinician Scientist Endowed Professorship from the Else Kr\"oner Fresenius Foundation (reference
number: 2022\_EKCS.17).

\bibliographystyle{unsrt}
\bibliography{references}

\clearpage
\appendix
\section*{Supplementary material}
\setcounter{figure}{0}
\setcounter{table}{0}
\renewcommand{\thefigure}{S\arabic{figure}}
\renewcommand{\thetable}{S\arabic{table}}
\renewcommand{\theHfigure}{S\arabic{figure}}
\renewcommand{\theHtable}{S\arabic{table}}

\begin{table}[h]
\centering
\caption{\textbf{Benchmark sources.} The build script downloads these sources and derives the
reference masks as defined in \texttt{benchmark/cases.json}.}
\label{tab:sources}
\footnotesize
\begin{tabular}{>{\raggedright\arraybackslash}p{2.9cm}>{\raggedright\arraybackslash}p{5.3cm}>{\raggedright\arraybackslash}p{6.1cm}}
\toprule
Image & Targets & Source, licence \\
\midrule
\texttt{pediatric\_ct} & liver, spleen, right kidney, bladder, heart, pancreas, gallbladder, left adrenal & Pediatric-CT-SEG (TCIA) \cite{jordan2021pediatricctseg}, CC BY 4.0 \\
\texttt{hcc\_ct} & liver, liver tumour & HCC-TACE-Seg (TCIA) \cite{moawad2021hcctaceseg}, CC BY 4.0 \\
\texttt{adrenal\_ct} & adrenal tumour & Adrenal-ACC-Ki67-Seg (TCIA) \cite{moawad2023adrenalacc}, CC BY 4.0 \\
\texttt{lung\_ct} & lung tumour & RIDER Lung CT, version 3 (TCIA) \cite{zhao2015riderlungct}, CC BY 4.0; tumour mask from RIDER-LungCT-Seg (TCIA) \cite{wee2020riderlungctseg}, CC BY 3.0 \\
\texttt{lung\_pet} & FDG-avid tumour & ACRIN-NSCLC-FDG-PET (TCIA) \cite{kinahan2019acrinnsclc}, CC BY 3.0; lesion mask from the BAMF/AIMI annotations \cite{vanoss2024bamf,murugesan2024aimi}, CC BY 4.0 \\
\texttt{headneck\_mr} & mandible, brainstem, left parotid, right submandibular gland, thyroid, spinal cord, left optic nerve & HaN-Seg \cite{podobnik2023hanseg}, CC BY-NC-ND 4.0 \\
\texttt{headneck\_tumor\_mr} & primary tumour & HNTS-MRG 2024 \cite{wahid2024hntsmrg}, CC BY-NC 4.0 \\
\texttt{bladder\_mr} & bladder tumour & FedBCa \cite{cao2024fedbca}, CC BY 4.0 \\
\texttt{thigh\_mr} & rectus femoris, sartorius, gracilis, vastus lateralis & Thigh and calf muscle MRI \cite{wohlfarth2025thighcalf}, on request from the authors (former figshare release: CC BY 4.0, non-commercial use only) \\
\texttt{calf\_mr} & soleus, tibialis anterior, medial gastrocnemius & same \\
\texttt{foot\_ct} & the 17 bones of the left foot: talus, calcaneus, navicular, cuboid, three cuneiforms, five metatarsals, the phalanges of each toe & Visible Human Full-Skeleton 3D Bone Models (Hugging Face) \cite{alavi2026visiblehuman}, CC BY 4.0, on the CT of the NLM Visible Human Female as aligned by \cite{andreassen2023visiblehuman}, CC BY 4.0; courtesy of the U.S. National Library of Medicine \\
\texttt{liver\_vessels\_ct} & liver, portal vein, hepatic veins & VEELA 2025 \cite{toprak2025veela,yazici2026veela}; images from CHAOS \cite{kavur2019chaosdata}, CC BY-NC-SA 4.0; no licence stated for the VEELA labels \\
\texttt{mouse\_microct} & subcutaneous tumour & Subcutaneous tumour micro-CT database (ERDA) \cite{jensen2024tumseg}, CC BY (version not stated) \\
\texttt{ant\_microct} & brain & Ant brains in micro-CT (Dryad) \cite{toulkeridou2023ants,toulkeridou2023antsdata}, CC0 1.0 \\
\bottomrule
\end{tabular}
\end{table}

\begin{figure}[h]
{\small\textbf{(a) Instructions}}
\begin{lstlisting}[style=promptstyle]
You segment images with the nninteractive tools. Before the first nninteractive tool call, read the nninteractive-segmentation skill and follow it. Work autonomously: there is no human to answer questions during the run.
\end{lstlisting}
{\small\textbf{(b) Guided request} (\texttt{adrenal\_ct})}
\begin{lstlisting}[style=promptstyle]
This is an abdominal CT scan: <path>/adrenal_ct/image.nii.gz.
Segment each of the following structures as its own object, named exactly by its id. For each, a voxel inside it is given ([axis0, axis1, axis2], the coordinates your tools use).
Work through them in this order and commit each object before you start the next one:
- t01_adrenal_tumor: the tumor of the right adrenal gland, contains voxel [74, 295, 165]
Save the result when you are done.
\end{lstlisting}
{\small\textbf{(c) Unguided request} (\texttt{headneck\_mr})}
\begin{lstlisting}[style=promptstyle]
This is a T1-weighted MRI of the head and neck: <path>/headneck_mr/image.nii.gz.
Segment each of the following structures as its own object, named exactly by its id.
Work through them in this order and commit each object before you start the next one:
- t01_mandible: the mandible
- t02_brainstem: the brainstem
- t03_left_parotid_gland: the left parotid gland
- t04_right_submandibular_gland: the right submandibular gland
- t05_thyroid_gland: the thyroid gland
- t06_spinal_cord: the spinal cord
- t07_left_optic_nerve: the left optic nerve
Save the result when you are done.
\end{lstlisting}
{\small\textbf{(d) VoxTell's prompts} (\texttt{headneck\_mr}, one call)}
\begin{lstlisting}[style=promptstyle]
"the mandible" "the brainstem" "the left parotid gland" "the right submandibular gland"
"the thyroid gland" "the spinal cord" "the left optic nerve"
\end{lstlisting}
\vspace{-4pt}
\caption{\textbf{What the agents and VoxTell receive} (verbatim; file paths shortened). Every request
follows one template: a one-line description of the image with its path, then the targets in a fixed
order, each with an id and a text, and in the guided track one voxel inside each target. Codex
receives the instructions (a) at the start of the message, Claude Code as an appended system prompt
(with ``load the nninteractive:'' in place of ``read''). VoxTell receives only the target texts.
The description and target texts of every image are listed in \texttt{benchmark/cases.json}.}
\label{fig:requests}
\end{figure}

\begin{figure}[h]
\centering
\includegraphics[width=\textwidth]{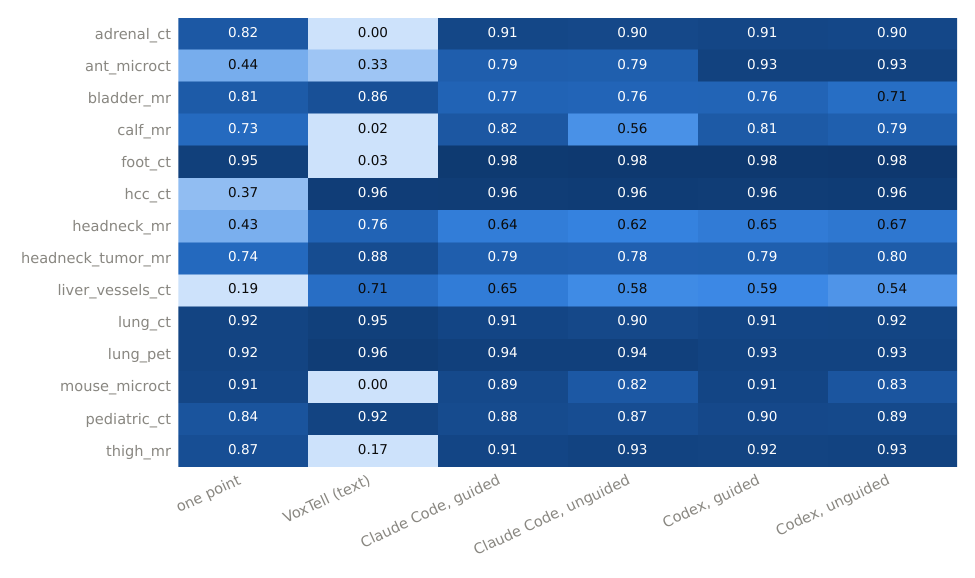}
\caption{\textbf{Mean Dice per image and track} (over the image's targets and, for the agents, the
three repeats).}
\label{fig:perimage}
\end{figure}

\begin{table}[p]
\centering
\caption{\textbf{Dice per target and track} (agents and one point: mean over three repeats; oracles:
after five prompts).}
\label{tab:pertarget}
\scriptsize
\setlength{\tabcolsep}{3.5pt}
\begin{tabular}{llcccccccc}
\toprule
Image & Target & \begin{tabular}[b]{@{}c@{}}Codex,\\guided\end{tabular} & \begin{tabular}[b]{@{}c@{}}Claude Code,\\guided\end{tabular} & \begin{tabular}[b]{@{}c@{}}Codex,\\unguided\end{tabular} & \begin{tabular}[b]{@{}c@{}}Claude Code,\\unguided\end{tabular} & \begin{tabular}[b]{@{}c@{}}One point\end{tabular} & \begin{tabular}[b]{@{}c@{}}VoxTell\end{tabular} & \begin{tabular}[b]{@{}c@{}}Scribble oracle,\\5 prompts\end{tabular} & \begin{tabular}[b]{@{}c@{}}Point oracle,\\5 prompts\end{tabular} \\
\midrule
\texttt{adrenal\_ct} & adrenal tumour & 0.91 & 0.91 & 0.90 & 0.90 & 0.82 & 0.00$^*$ & 0.93 & 0.91 \\
\midrule
\texttt{ant\_microct} & brain & 0.93 & 0.79 & 0.93 & 0.79 & 0.44 & 0.33 & 0.94 & 0.74 \\
\midrule
\texttt{bladder\_mr} & bladder tumour & 0.76 & 0.77 & 0.71 & 0.76 & 0.81 & 0.86 & 0.86 & 0.85 \\
\midrule
\texttt{calf\_mr} & soleus & 0.75 & 0.79 & 0.74 & 0.58 & 0.57 & 0.00 & 0.88 & 0.84 \\
 & tibialis anterior & 0.83 & 0.81 & 0.82 & 0.63 & 0.72 & 0.06 & 0.91 & 0.86 \\
 & medial gastrocnemius & 0.86 & 0.87 & 0.80 & 0.46 & 0.90 & 0.00 & 0.95 & 0.94 \\
\midrule
\texttt{foot\_ct} & left talus & 0.99 & 1.00 & 0.99 & 0.99 & 1.00 & 0.38 & 0.99 & 0.99 \\
 & left calcaneus & 0.99 & 0.99 & 0.99 & 0.99 & 0.99 & 0.06 & 0.99 & 0.99 \\
 & left navicular & 0.99 & 0.99 & 0.99 & 0.99 & 1.00 & 0.00 & 0.99 & 0.99 \\
 & left cuboid & 0.99 & 0.99 & 0.99 & 0.99 & 0.99 & 0.00 & 0.99 & 0.99 \\
 & left medial cuneiform & 0.99 & 0.99 & 0.99 & 0.99 & 0.99 & 0.00 & 0.99 & 0.99 \\
 & left intermediate cuneiform & 0.98 & 0.98 & 0.98 & 0.98 & 0.99 & 0.00 & 0.98 & 0.98 \\
 & left lateral cuneiform & 0.99 & 0.99 & 0.99 & 0.99 & 0.99 & 0.00 & 0.98 & 0.99 \\
 & left metatarsal 1 & 0.99 & 0.99 & 0.99 & 0.99 & 0.92 & 0.00 & 0.99 & 0.99 \\
 & left metatarsal 2 & 0.97 & 0.98 & 0.98 & 0.98 & 0.99 & 0.00 & 0.98 & 0.99 \\
 & left metatarsal 3 & 0.99 & 0.98 & 0.99 & 0.98 & 0.99 & 0.00 & 0.98 & 0.98 \\
 & left metatarsal 4 & 0.98 & 0.98 & 0.99 & 0.98 & 0.99 & 0.00 & 0.97 & 0.98 \\
 & left metatarsal 5 & 0.99 & 0.98 & 0.98 & 0.98 & 0.91 & 0.00 & 0.99 & 0.99 \\
 & left big toe phalanges & 0.97 & 0.97 & 0.97 & 0.98 & 0.78 & 0.00 & 0.98 & 0.99 \\
 & left toe 2 phalanges & 0.96 & 0.97 & 0.97 & 0.96 & 0.88 & 0.03 & 0.98 & 0.98 \\
 & left toe 3 phalanges & 0.92 & 0.92 & 0.98 & 0.94 & 0.87 & 0.00 & 0.96 & 0.98 \\
 & left toe 4 phalanges & 0.97 & 0.96 & 0.97 & 0.96 & 0.86 & 0.00 & 0.98 & 0.97 \\
 & left toe 5 phalanges & 0.97 & 0.95 & 0.97 & 0.96 & 0.94 & 0.00 & 0.98 & 0.98 \\
\midrule
\texttt{hcc\_ct} & liver & 0.97 & 0.96 & 0.97 & 0.97 & 0.61 & 0.97 & 0.88 & 0.97 \\
 & liver tumour & 0.96 & 0.96 & 0.96 & 0.96 & 0.13 & 0.94 & 0.96 & 0.97 \\
\midrule
\texttt{headneck\_mr} & mandible & 0.74 & 0.70 & 0.77 & 0.73 & 0.20 & 0.81 & 0.80 & 0.82 \\
 & brainstem & 0.77 & 0.81 & 0.80 & 0.75 & 0.27 & 0.86 & 0.82 & 0.81 \\
 & left parotid gland & 0.82 & 0.83 & 0.81 & 0.83 & 0.83 & 0.87 & 0.90 & 0.89 \\
 & right submandibular gland & 0.80 & 0.79 & 0.80 & 0.80 & 0.80 & 0.85 & 0.89 & 0.85 \\
 & thyroid gland & 0.68 & 0.68 & 0.66 & 0.66 & 0.48 & 0.76 & 0.77 & 0.72 \\
 & spinal cord & 0.49 & 0.46 & 0.52 & 0.55 & 0.41 & 0.61 & 0.79 & 0.57 \\
 & left optic nerve & 0.27 & 0.22 & 0.34 & 0.02 & 0.02 & 0.59 & 0.64 & 0.46 \\
\midrule
\texttt{headneck\_tumor\_mr} & primary tumour & 0.79 & 0.79 & 0.80 & 0.78 & 0.74 & 0.88 & 0.88 & 0.88 \\
\midrule
\texttt{liver\_vessels\_ct} & liver & 0.95 & 0.96 & 0.95 & 0.96 & 0.00 & 0.96 & 0.96 & 0.95 \\
 & portal vein & 0.41 & 0.47 & 0.42 & 0.50 & 0.35 & 0.53 & 0.49 & 0.47 \\
 & hepatic veins & 0.42 & 0.53 & 0.27 & 0.30 & 0.20 & 0.64 & 0.44 & 0.53 \\
\midrule
\texttt{lung\_ct} & lung tumour & 0.91 & 0.91 & 0.92 & 0.90 & 0.92 & 0.95 & 0.93 & 0.94 \\
\midrule
\texttt{lung\_pet} & tumour & 0.93 & 0.94 & 0.93 & 0.94 & 0.92 & 0.96 & 0.94 & 0.94 \\
\midrule
\texttt{mouse\_microct} & tumour & 0.91 & 0.89 & 0.83 & 0.82 & 0.91 & 0.00 & 0.96 & 0.96 \\
\midrule
\texttt{pediatric\_ct} & liver & 0.98 & 0.98 & 0.98 & 0.98 & 0.98 & 0.98 & 0.98 & 0.98 \\
 & spleen & 0.98 & 0.98 & 0.98 & 0.98 & 0.98 & 0.98 & 0.98 & 0.98 \\
 & right kidney & 0.98 & 0.97 & 0.98 & 0.97 & 0.97 & 0.98 & 0.98 & 0.98 \\
 & urinary bladder & 0.96 & 0.96 & 0.96 & 0.96 & 0.97 & 0.96 & 0.98 & 0.98 \\
 & heart & 0.93 & 0.92 & 0.91 & 0.90 & 0.42 & 0.96 & 0.93 & 0.93 \\
 & pancreas & 0.81 & 0.76 & 0.73 & 0.68 & 0.89 & 0.91 & 0.92 & 0.91 \\
 & gallbladder & 0.96 & 0.96 & 0.96 & 0.96 & 0.95 & 0.96 & 0.96 & 0.96 \\
 & left adrenal gland & 0.58 & 0.53 & 0.65 & 0.54 & 0.55 & 0.66 & 0.75 & 0.68 \\
\midrule
\texttt{thigh\_mr} & rectus femoris & 0.95 & 0.95 & 0.94 & 0.94 & 0.95 & 0.00 & 0.96 & 0.96 \\
 & sartorius & 0.93 & 0.90 & 0.92 & 0.92 & 0.91 & 0.27 & 0.94 & 0.94 \\
 & gracilis & 0.91 & 0.93 & 0.92 & 0.93 & 0.92 & 0.41 & 0.94 & 0.93 \\
 & vastus lateralis & 0.90 & 0.87 & 0.92 & 0.91 & 0.70 & 0.00 & 0.95 & 0.94 \\
\bottomrule
\multicolumn{10}{l}{$^*$ With the prompt ``adrenal tumor'', VoxTell reaches Dice 0.79 (a check outside the benchmark).} \\
\end{tabular}

\end{table}

\begin{figure}[p]
\centering
\includegraphics[width=\textwidth]{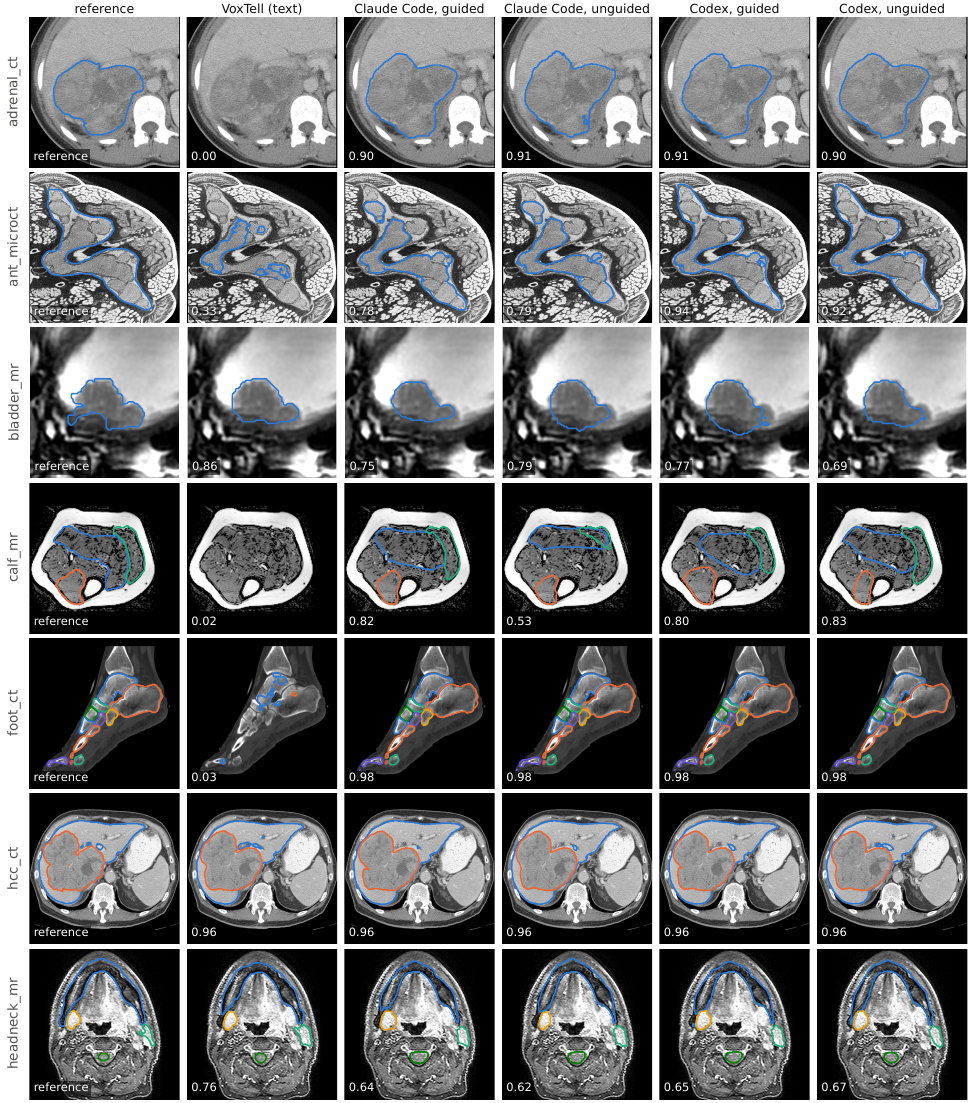}
\caption{\textbf{All tracks}, images 1--7: reference, VoxTell, and the middle of three repeats of
each agent track. Numbers: Dice of that run, averaged over the image's targets.}
\label{fig:casesall1}
\end{figure}

\begin{figure}[p]
\centering
\includegraphics[width=\textwidth]{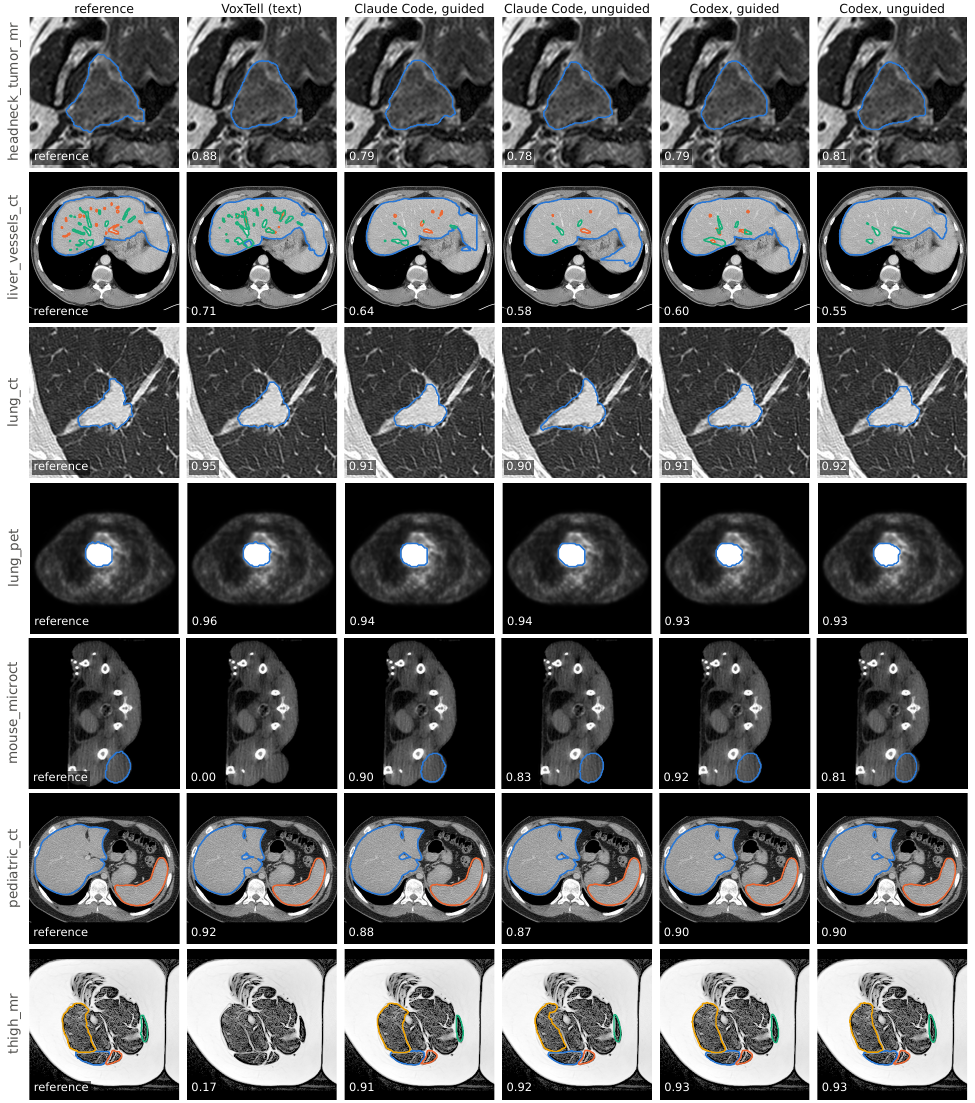}
\caption{\textbf{All tracks}, images 8--14 (as Fig.~\ref{fig:casesall1}).}
\label{fig:casesall2}
\end{figure}

\begin{figure}[p]
\centering
\includegraphics[width=\textwidth]{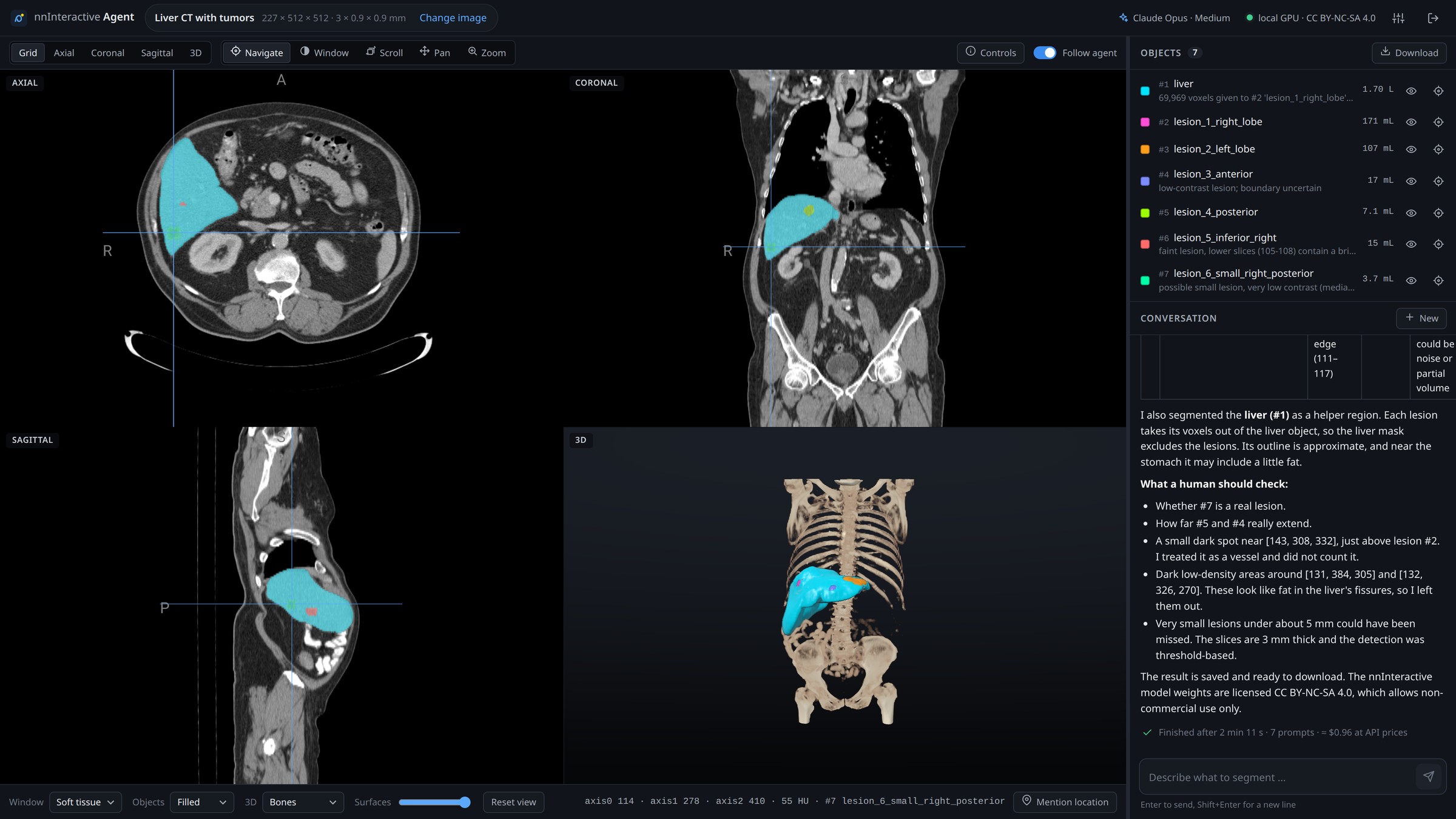}
\caption{\textbf{The web demo} after ``Segment and count all liver lesions.'' (MSD Liver, case 51,
Claude Opus 5.5): axial, coronal, sagittal and 3D views, the objects, and the
agent's report.}
\label{fig:webdemo}
\end{figure}

\begin{figure}[p]
\centering
\includegraphics[width=\textwidth]{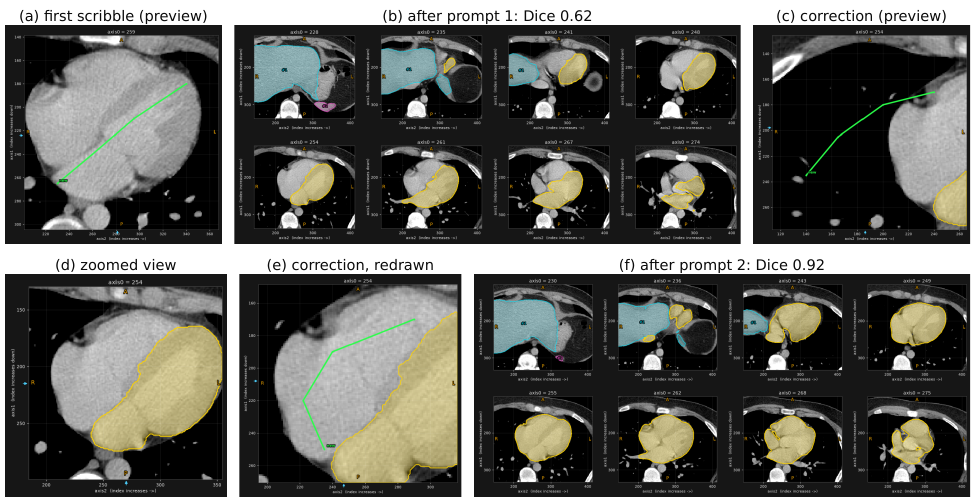}
\caption{\textbf{Previews and slices}: Claude Code segmenting the heart in a paediatric CT
(\texttt{pediatric\_ct}, Pediatric-CT-SEG \cite{jordan2021pediatricctseg}; guided track). All
panels are renders the agent requested. (a) The preview of its first scribble. (b) A montage after
this prompt: only the left chambers are segmented (Dice 0.62). (c) The preview of a correction: the
line lies in the lung, and the tool reports air intensities along it. The agent dropped this line
without sending it. (d) A zoomed view of the slice: the mask covers the left chambers but not the
right heart. The agent wrote ``Heart prediction only covers the left chambers; adding the right
heart with a positive scribble'' and (e) drew the line again, inside the right heart, and sent it.
(f) The montage after this second prompt. The agent committed the heart (Dice 0.92) with the note
``Includes proximal pulmonary trunk root; great vessels otherwise excluded.''}
\label{fig:sup_heart}
\end{figure}

\begin{figure}[p]
\centering
\includegraphics[width=\textwidth]{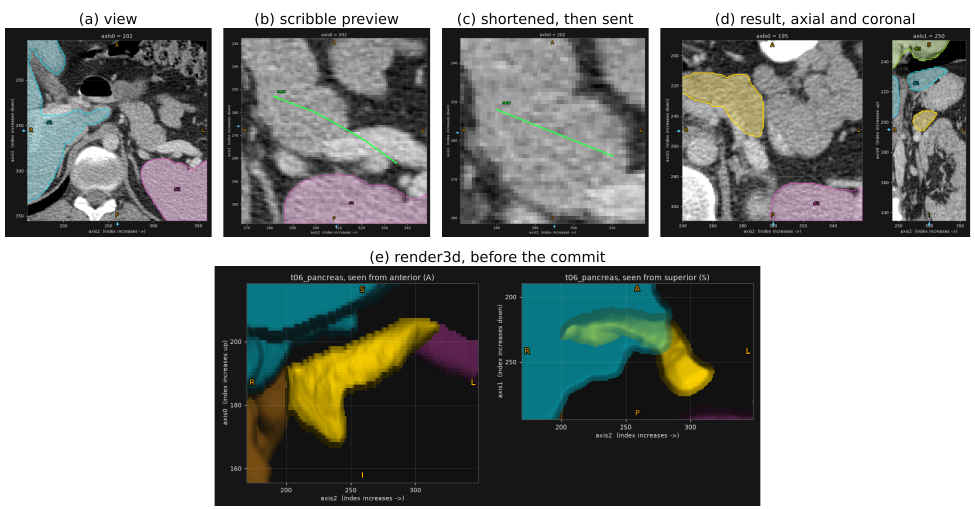}
\caption{\textbf{A single prompt}: Claude Code segmenting the pancreas, a low-contrast organ, in a
paediatric CT (\texttt{pediatric\_ct}; guided track). (a) The view in which the agent looked for
the pancreas, with voxel coordinates, patient directions and the organs committed so far. (b) The
preview of its first scribble. (c) The agent shortened the scribble and sent it. (d) The result of
this single prompt, checked in an axial and a coronal slice. (e) The 3D rendering of the pancreas
(yellow) with its committed neighbours. The agent then committed it (Dice 0.89) with the note
``Low contrast against adjacent bowel; tail and head boundaries uncertain.''}
\label{fig:sup_pancreas}
\end{figure}

\begin{figure}[p]
\centering
\includegraphics[width=\textwidth]{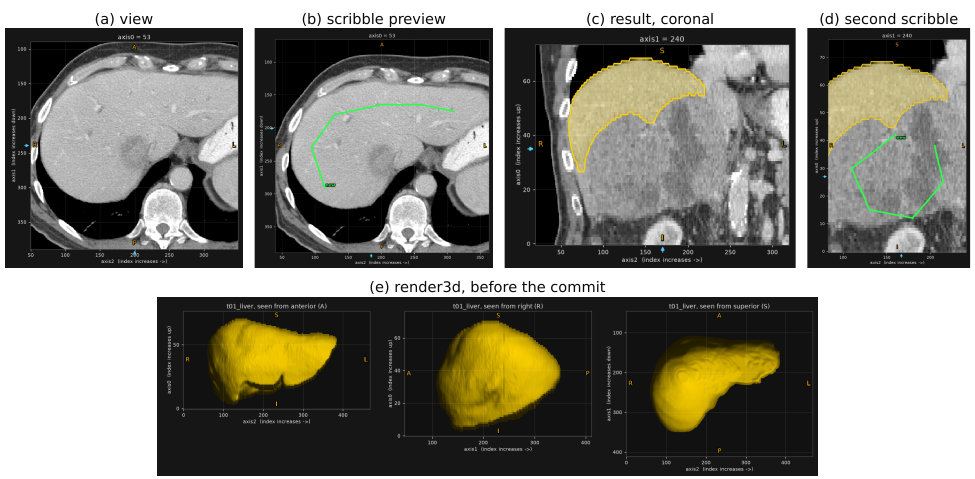}
\caption{\textbf{A check in another plane}: Claude Code segmenting the liver in a contrast-enhanced
CT from the request alone (\texttt{hcc\_ct}, HCC-TACE-Seg \cite{moawad2021hcctaceseg}; unguided
track). (a) An axial view with voxel coordinates and patient directions. (b) The preview of its
first scribble, along the liver. (c) The result after that prompt (Dice 0.66), checked in a coronal
slice: the large tumour is missing (the agent segments the liver including its tumour and separates
the tumour later). (d) The preview of a second scribble in that slice, which adds the tumour.
(e) The 3D rendering of the liver before the agent commits it (Dice 0.97).}
\label{fig:sup_liver}
\end{figure}

\begin{figure}[p]
\centering
\includegraphics[width=\textwidth]{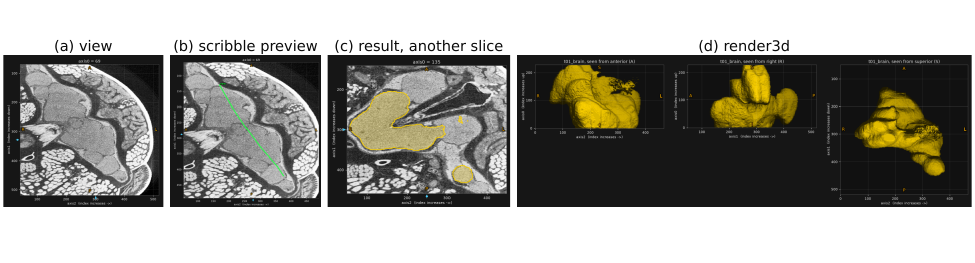}
\caption{\textbf{An insect brain}: Claude Code segmenting the brain in a micro-CT of an ant's head
(\texttt{ant\_microct} \cite{toulkeridou2023ants,toulkeridou2023antsdata}; guided track). (a) A
view with voxel coordinates and anatomical directions. (b) The preview of its first scribble, along
the brain. (c) The result after that prompt, checked in another slice. (d) The 3D rendering of the
object from three directions. The final object, after four prompts (one of them undone), reaches
Dice 0.82.}
\label{fig:sup_ant}
\end{figure}

\end{document}